\documentclass[10pt,twocolumn,letterpaper]{article}

\usepackage[pagenumbers]{genericconf}   
\usepackage{amsmath,amsthm}

\newtheorem{theorem}{Theorem}

\usepackage{bm}
\usepackage{algorithm}
\usepackage{algorithmic}
\usepackage{multirow}
\usepackage[dvipsnames]{xcolor}
\usepackage{listings}
\usepackage{mathtools}
\lstdefinestyle{pythonstyle}{
  language=Python,
  basicstyle=\ttfamily\footnotesize,
  keywordstyle=\color{RoyalBlue}\bfseries,
  commentstyle=\color{ForestGreen},
  stringstyle=\color{BrickRed},
  numberstyle=\tiny\color{gray},
  numbers=left,
  stepnumber=1,
  showstringspaces=false,
  breaklines=true,
  frame=single,
  framerule=0.3pt,
  tabsize=4
}

\usepackage{booktabs}
\usepackage[pagebackref,breaklinks,colorlinks]{hyperref}

\title{Shortcut Invariance: Targeted Jacobian Regularization \\ in Disentangled Latent Space}

\author{Shivam Pal\\
IIT Kanpur\\
{\tt\small pshivam@cse.iitk.ac.in}
\and
Sakshi Varshney\\
ARF, IIT Kanpur\\
{\tt\small sakshi.varshney@airawat.org}
\and 
Piyush Rai\\
IIT Kanpur \\
{\tt\small piyush@cse.iitk.ac.in}
}
\begin{document}
\maketitle

\begin{abstract}
Deep neural networks are prone to learning \emph{shortcuts}, spurious correlations 
present in the training data that undermine out-of-distribution (OOD) generalization. 
Most prior work mitigates shortcut learning through input-space reweighting, either 
relying on explicit shortcut labels or inferring shortcut structure from heuristics 
such as per-sample loss. Moreover, these approaches typically assume the presence of some
\emph{shortcut-conflicting} examples in the training set, an assumption that is often violated in practice, particularly in medical imaging where data is aggregated across institutions with different acquisition protocols.

We propose a latent-space method that views shortcut learning as over-reliance on shortcut-aligned axes. In a disentangled latent space, we identify candidate shortcut-aligned axes via their strong correlation with labels and reduce classifier reliance on them by injecting targeted anisotropic noise during training. Unlike prior latent-space based approaches that remove, project out, or adversarially suppress shortcut features, our method preserves the full representation and instead impose functional invariance by regularizing the classifier’s sensitivity along those axes.

We show that injecting anisotropic noise induces targeted Jacobian and curvature regularization, effectively flattening the decision boundary along shortcut axes while leaving core feature dimensions largely unaffected. Our method achieves state-of-the-art OOD performance across standard shortcut-learning benchmarks without requiring shortcut labels or shortcut-conflicting samples.
\end{abstract}    
\section{Introduction}
\label{sec:intro}
\begin{figure*}
    \centering
    \includegraphics[width=1.0\linewidth]{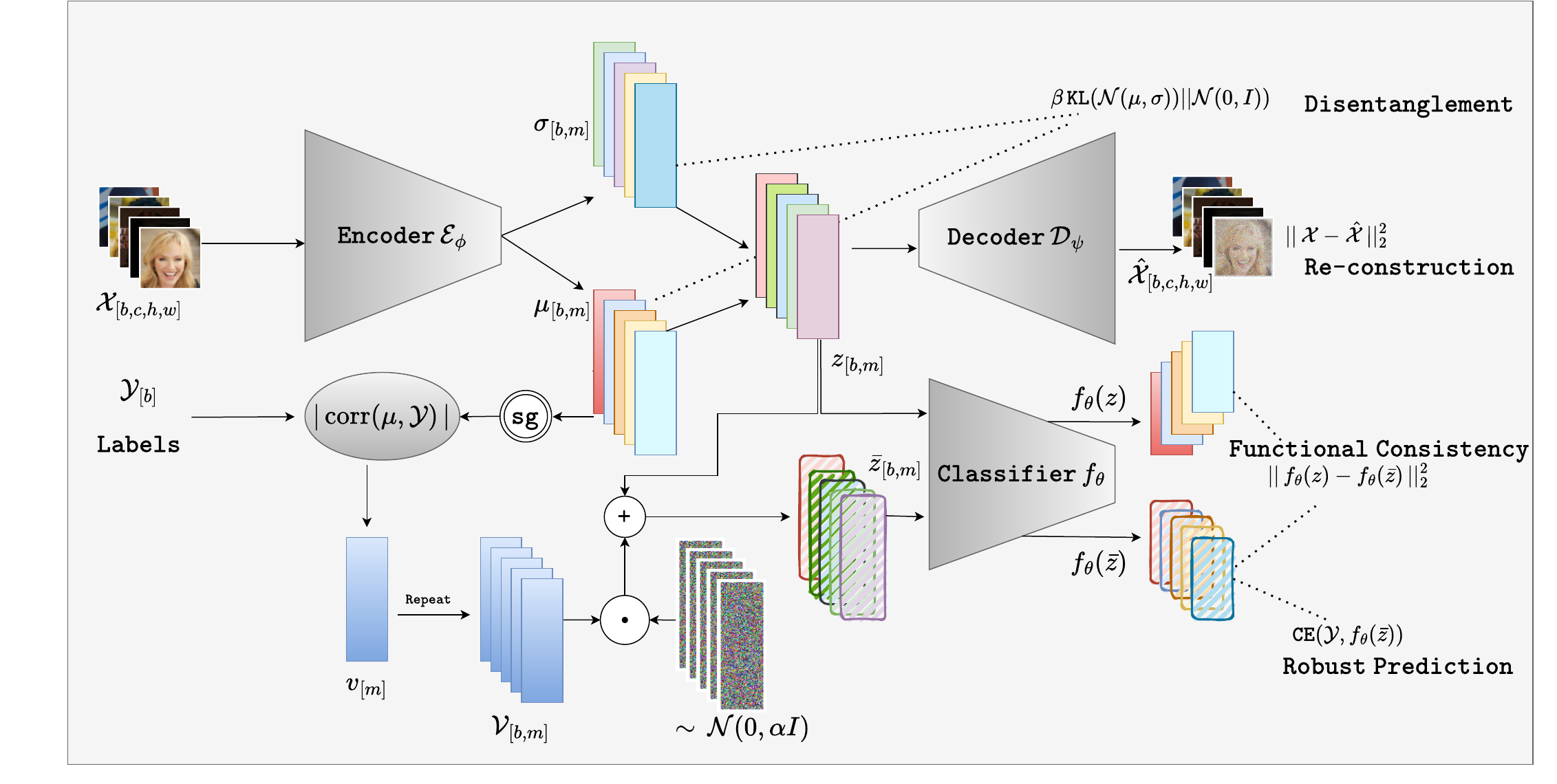}
    \caption{\textbf{Overview of SITAR.} A $\beta$-VAE encoder $\mathcal{E}_\phi$ maps input
    images $\mathcal{X}$ to Gaussian latents
    $\bm{z} \sim \mathcal{N}(\bm{\mu},\bm{\sigma})$, which are then passed to a decoder
    $\mathcal{D}_\psi$ for reconstruction and to a classifier $f_\theta$ for prediction.
    Using labels $\mathcal{Y}$ and latent means $\bm{\mu}$, SITAR computes
    per-dimension shortcut scores
    $v_j = \lvert \operatorname{corr}(\mu_j,\mathcal{Y}) \rvert$ (with gradients
    stopped), forming a weight vector $\bm{v}$. Independent Gaussian noise
    $\bm{\epsilon} \sim \mathcal{N}(0,\alpha I)$ is scaled elementwise by
    $\bm{v}$ and added to $\bm{z}$ to obtain perturbed latents $\bar{\bm{z}}$.
    The encoder $E_\phi$, decoder $D_\psi$, and classifier $f_\theta$ are trained
    \emph{jointly} using the sum of four losses: reconstruction, $\beta$-weighted
    KL divergence, cross-entropy on $(\bar{\bm{z}},\mathcal{Y})$, and an $\ell_2$
    consistency loss $\lVert f_\theta(\bm{z}) - f_\theta(\bar{\bm{z}})\rVert_2^2$,
    encouraging shortcut-invariant decision functions in the disentangled latent
    space.}

    \label{fig:sitar}
\end{figure*}
Deep neural networks trained with Empirical Risk Minimization (ERM) \cite{vapnik1995} achieve superhuman performance on in-distribution test sets, yet systematically fail under distribution shift.
A central cause of this failure is \textit{shortcut learning}, networks preferentially learn low-complexity spurious correlations, that are predictive in the training distribution but break under shift, rather than the intended semantic rules~\cite{geirhos2020shortcut}. For instance, digit classifiers in CMNIST \cite{arjovsky2019invariant} exploit background color rather than shape, and texture-biased ImageNet models \cite{geirhos2018imagenet} resolve object identity by texture rather than form. In both cases, the spurious feature is more predictive than the causal one during training, so ERM relies on spurious features rather than the intended core features.

 Most prior work addresses shortcut learning by reweighting training samples to reduce reliance on spurious features, either using explicit shortcut group labels \cite{arjovsky2019invariant, sagawa2019distributionally} or inferring which samples are shortcut-reliant from proxies such as per-sample loss \cite{nam2020learningfailuretrainingdebiased, levy2020large, liu2021justtraintwiceimproving}. However, these approaches share a critical assumption: the training set must contain \emph{shortcut-conflicting}~\cite{nam2020learningfailuretrainingdebiased} examples, i.e., samples where the spurious feature is absent or misleading. In practice, this is often not the case. In medical imaging, for instance, spurious correlations arise systematically when data is aggregated across institutions with different acquisition protocols~\cite{degrave2021}, leaving no conflicting examples to exploit.

An alternative line of work addresses shortcut learning at the representation level, aiming to partition the features space into a \textit{core} component encoding causal features and a \textit{spurious} component encoding shortcut features, with the downstream classifier trained exclusively on the former. Such methods, however, carry significant limitations, they often require explicit labels for the spurious attribute~\cite{shortcut2023Nicolas}, rely on separability assumptions between core and spurious features that rarely hold, or fail to scale when the shortcut signal is high-dimensional or entangled with semantic content~\cite{yang2022chroma}.

In this paper, rather than enforcing shortcut invariance through representational purity, we ask a different question: can we train a classifier that is \textit{functionally invariant} to shortcut signals, regardless of whether they are present in its input representation? 
We propose \textbf{SITAR} (\textbf{S}hortcut \textbf{I}nvariance via \textbf{T}argeted \textbf{A}nisotropic \textbf{R}egularization), a method that trains a classifier to be \textit{functionally invariant} to shortcuts without requiring shortcut labels or a shortcut-free representation.

\textbf{Core Hypothesis:} In a disentangled representation, shortcut features occupy latent dimensions that exhibit a \textit{stronger correlation} with the labels than dimensions encoding  core features.

This correlation gap serves as an effective, unsupervised proxy for identifying shortcut dimensions, requiring no shortcut labels. 
We exploit it directly by injecting anisotropic noise into the latent vector during training, applying the strongest perturbations to dimensions most correlated with the label.
The classifier $f_\theta$ is trained with a consistency-based objective. it must produce a correct prediction on the perturbed latent vector $\bar{\bm{z}}$ while remaining consistent with its output on the original $\bm{z}$, i.e., $f_\theta(\bm{z}) \approx f_\theta(\bar{\bm{z}})$.
Since the strongest noise is applied to the shortcut dimensions, this objective directly penalizes reliance on them, pushing the classifier to derive its predictions from the low-correlation dimensions that encode core semantic content. Crucially, the consistency objective does not suppress the highly correlated dimensions entirely. It merely desensitizes the classifier to them. As a result, when no shortcuts are present in the data, the classifier still has access to all predictive signal and performance does not degrade, which we verify empirically.

Our theoretical analysis formally justifies the SITAR objective Eq.~\eqref{eq:obj}. A second-order small-noise expansion shows that this is analytically equivalent to augmenting the ERM loss with a \textit{targeted Jacobian} and \textit{curvature regularizer}. Dimensions with high correlation (shortcuts) are penalized most strongly, directly flattening the classifier along those dimensions and suppressing its sensitivity to shortcut features.

Jacobian regularization \cite{DBLP:journals/corr/abs-1908-02729} and curvature regularization \cite{Moosavi-Dezfooli_2019_CVPR} are established techniques for improving the robustness of classifiers, typically applied uniformly across all input dimensions. Theorem~\ref{thm:sitar_expansion} reveals that SITAR implicitly performs both, but critically in a \textit{non-uniform} manner. This anisotropy is the key mechanism that uniform regularization cannot achieve and that, to our knowledge, has not been previously exploited for shortcut mitigation.

SITAR is simple to implement, requires no shortcut labels, and is empirically robust across the full spectrum of shortcut severity,  from settings with no shortcuts to those where no shortcut-conflicting examples exist.  Our contributions are as follows:
\begin{enumerate}
\item A training method that enforces \textit{functional invariance} to shortcut signals at the classifier level, without requiring shortcut labels.
\item A theoretical analysis showing that our consistency objective is equivalent to a targeted Jacobian and Curvature regularizer that penalizes classifier sensitivity along shortcut dimensions, weighted by their correlation strength.
\item Extensive experiments on real-world benchmark datasets demonstrate state-of-the-art worst-group accuracy. Notably, our method remains robust even when shortcut-conflicting examples are absent from the training data, a regime where prior methods typically fail.

\end{enumerate}
\section{Related Work}
\label{sec:related}

Deep neural networks often rely on \emph{shortcuts}—spurious yet predictive cues that enable easy optimization but poor out-of-distribution (OOD) generalization~\cite{geirhos2020shortcut}. Optimization- and data-level debiasing methods such as Group DRO~\cite{sagawa2019distributionally}, IRM~\cite{arjovsky2019invariant}, REx~\cite{krueger2021out}, JTT~\cite{liu2021justtraintwiceimproving}, and Learning from Failure~\cite{nam2020learningfailuretrainingdebiased} modify training dynamics to reduce shortcut reliance. More recent adversarial and augmentation based strategies including BiasAdv~\cite{lim2023biasadv}, learnable data perturbations~\cite{morerio2024learnable}, and counterfactual data generation~\cite{chang2021robust}—further aim to neutralize spurious cues. However, these approaches typically depend on group or pseudo-group labels, involve multi-stage or adversarial pipelines, and are sensitive to hyperparameters, often resulting in substantial computational overhead.

\textbf{Disentanglement-based} generative models such as $\beta$-VAE~\cite{higgins2017beta} and FactorVAE~\cite{kim2018disentangling} separate independent latent factors, providing interpretability but not necessarily invariance. Recent works~\cite{muller2023shortcutvae,yang2022chroma} connect disentanglement with shortcut detection or mitigation. ShortcutVAE~\cite{muller2023shortcutvae} reveals shortcut-sensitive latent dimensions but relies on human inspection to remove them. 
Chroma-VAE~\cite{yang2022chroma} mitigates shortcuts through coupled generative--discriminative training by heuristically partitioning the latent space into \textit{core} and \textit{shortcut} dimensions, discarding the latter and training a $k$-nearest-neighbor ($k$NN) classifier on the core latents. While it is claimed that any classifier can be used on this representation, we found that Chroma-VAE's robustness largely disappears when $k$NN is replaced by parametric models such as logistic regression or small neural networks. This gap arises because the shortcut factors are not fully removed from the core latents but only downscaled. Distance-based methods like $k$NN are highly sensitive to feature scaling, whereas linear and neural classifiers can effectively undo such rescaling through their learned weights. As a result, Chroma-VAE's invariance is tightly coupled to a particular non-parametric classifier, and the underlying representation still carries shortcut information.

\textbf{Generative modeling} provides an additional route for mitigating shortcuts.
 Diffusion-based models synthesize counterfactual or attribute-modified samples to break spurious correlations~\cite{scimeca2023,weng2024fastdiffusion}, while generative classifiers~\cite{li2025generative} mitigate shortcuts by modeling $p(x \mid y)$ to capture full class variability. However, these methods require computationally intensive generative modeling and prior knowledge of which attributes to modify. In contrast, our method remains purely discriminative: we use disentangled generative models only as latent feature learners and introduce an explicit shortcut-suppression loss, avoiding counterfactual synthesis or likelihood-based training while still mitigating shortcut reliance.

\section{Methodology}
\label{sec:method}

We now formally introduce \textbf{SITAR}, a method that trains a classifier $f_\theta$ that is \textit{functionally invariant} to shortcut features in a disentangled latent representation.
We present the method in terms of raw input $\bm{x}$, though a pretrained encoder can be used in place of pixel-space input (see Section~\ref{sec:pretrained}). The method proceeds in two steps, first identifying shortcut-aligned dimensions via label correlation (Section~\ref{sec:method_step1}), then desensitizing the classifier to those dimensions via targeted anisotropic noise (Section~\ref{sec:method_step2}).

\subsection{Disentangled Latent Representation}
\label{sec:method_pre}

Our approach operates on a disentangled latent representation $\bm{z} \in \mathbb{R}^m$ obtained via a $\beta$-VAE~\cite{higgins2017beta}. We adopt the standard ELBO objective:
\begin{align} 
\label{eq:vae-elbo}
\mathcal L_{\text{VAE}}(\bm{x})
&= \mathbb{E}_{q_\phi(\bm{z}\mid \bm{x})}[-\log p_\psi(\bm{x}\mid \bm{z})] \notag
\\[3pt] 
&\quad+\; \beta \; \mathrm{KL}\!\big(q_\phi(\bm{z}\mid \bm{x})\,\|\,p(\bm{z})\big),
\end{align}

 $q_\phi(\bm{z}\mid \bm{x})$ is the posterior distribution of $\bm{z}$ and it is assumed to be a Gaussian $\mathcal{N}(\bm{\mu(x)}, \bm{\sigma(x)})$ , where $\bm{\mu(x)}$ and $ \bm{\sigma(x)}$ are computed via a encoder network  $\mathcal{E}_{\phi}$. 
$p_\psi(\bm{x}\mid \bm{z})$ is computed via a deterministic decoder network $\mathcal{D}_{\psi}$. $p(\bm{z})$ is the prior distribution of $\bm z$ and it is assumed to be $\mathcal{N}(\bm{0},\bm I)$. $\mathrm{KL}(q \, ||\,p)$ is the Kullback-Leibler divergence between $q$ and $p$.  The hyperparameter $\beta > 1$ encourages disentanglement by imposing a stronger penalty on posterior deviation from the prior.
For a input $\bm x$, we denote its posterior mean and sampled latent by
 $\bm \mu$ and $\bm z$ respectively.

\subsection{Identifying Shortcut Proxies}
\label{sec:method_step1}
We use label correlation as an unsupervised proxy for identifying shortcut-aligned dimensions. Given a training set $\{(\bm{x}_n,y_n)\}_{n=1}^N$,  we first compute the mean $\bm{\mu}_n$ of the latent $\bm{z}_n$ for each input $\bm{x}_n$.
Given $\{(\bm{\mu}_n, y_n)\}_{n=1}^{N}$, where $\bm{\mu}_n \in \mathbb{R}^m$ and $y_n \in \mathbb{R}$, we define a vector $\bm{v} \in \mathbb{R}^m$ as follows. For each $j = 1, \dots, m$, let
$
\bm{\mu}^{(j)} := (\mu_{1j}, \dots, \mu_{Nj})^\top \in \mathbb{R}^N,
$
$
\bm{y} := (y_1, \dots, y_N)^\top \in \mathbb{R}^N.
$
Then the $j$-th component of $\bm{v}$ is
\begin{equation}
\label{eq:corr}
v_j := \bigl|\operatorname{Corr}(\bm{\mu}^{(j)}, \bm{y})\bigr|, \qquad j = 1, \dots, m,
\end{equation}

where $$ \operatorname{Corr}(\bm{a}, \bm{b})
= \frac{\operatorname{Cov}(\bm{a}, \bm{b})}
       {\sqrt{\operatorname{Var}(\bm{a})}\,\sqrt{\operatorname{Var}(\bm{b})}} $$
\noindent A high value of $v_j$ indicates that dimension $j$ is strongly aligned with the label and is therefore a candidate shortcut dimension. $\bm{v}$ forms the basis of our targeted regularization in the next section.

\subsection{Functional Invariance via Anisotropic Regularization}
\label{sec:method_step2}
 Having identified the shortcut sensitivity vector $\bm{v}$, we now define the training objective that enforces functional invariance to shortcut dimensions. We construct a perturbed latent vector $\bar{\bm{z}}$ by injecting anisotropic Gaussian noise scaled by $\bm{v}$,
\begin{equation}
\label{eq:zbar}
\bar{\bm{z}} = \bm{z} + \alpha \cdot (\bm{v} \odot \bm{e}), \qquad \bm{e} \sim \mathcal{N}(\bm{0}, \mathbf{I}),
\end{equation}
where $\odot$ denotes the Hadamard product and $\alpha > 0$ controls the overall perturbation strength. Dimensions with high $v_j$ (shortcut axes) receive high-variance noise, while dimensions with low $v_j$ (core axes) remain largely unperturbed.

The classifier $f_\theta$ is trained with a two-term objective. The \textbf{Robust Prediction} term applies standard cross-entropy on the perturbed input $\bar{\bm{z}}$, pushing $f_\theta$ to find predictive signal in the core axes, which are unaffected by the noise. The \textbf{Functional Consistency} term penalizes the change in classifier output under the perturbation, directly regularizing its sensitivity along the shortcut axes. Together, these yield the full training objective, optimized jointly with the VAE,
\begin{align}
\label{eq:obj}
\mathcal{L}(\bm{x}, y)
=\; \mathcal{L}_{\text{VAE}}(\bm{x})
+\; \underbrace{\mathbb{E}_{\bm{e}}\!\left[\ell_{\text{CE}}\!\big(f_\theta(\bar{\bm{z}}), y\big)\right]}_{\text{Robust Prediction}} \nonumber\\
+\; \lambda_{\text{cons}}
\underbrace{\mathbb{E}_{\bm{e}}\!\left[\|f_\theta(\bm{z}) - f_\theta(\bar{\bm{z}})\|_2^2\right]}_{\text{Functional Consistency}}.
\end{align}
As we show in Section~\ref{sec:theory}, both terms together are analytically equivalent to augmenting the ERM loss with targeted Jacobian and curvature regularizers, each weighted by $v_j^2$, that flatten the classifier along shortcut axes. The expectations in Eq.~\eqref{eq:obj} are approximated via a single-sample Monte Carlo estimate at each forward pass, providing an unbiased stochastic gradient for both terms. The shortcut sensitivity vector $\bm{v}$ is computed batchwise from detached latent means $\texttt{sg}(\bm{\mu})$. The VAE encoder $\mathcal{E}_\phi$ is updated by both $\mathcal{L}_{\text{VAE}}$ and $\mathcal{L}_{\text{clf}}$, as gradients propagate back through $f_\theta(\bm{z})$ and $f_\theta(\bar{\bm{z}})$, allowing the encoder to be jointly optimized with the classifier.

At test time, prediction uses the posterior mean directly,
\begin{equation}
\hat{y}(\bm{x}) = \arg\max_{y \in \mathcal{Y}} \left[f_\theta(\bm{\mu}(\bm{x}))\right]_y.
\end{equation}

\begin{algorithm}[t]
\small
\caption{SITAR: Shortcut Invariance via Targeted Anisotropic Regularization}
\label{alg:sitar}
\begin{algorithmic}[1]
\STATE \textbf{Input:} training set $D_{\mathrm{tr}}=\{(\bm{x}_n,\bm{y}_n)\}$; encoder $\mathcal{E}_{\phi}$, decoder $\mathcal{D}_{\psi}$; classifier $f_{\theta}$; hyperparams $\alpha,\beta,\lambda$
\STATE \textbf{Output:} $\mathcal{E}_{\phi}$, $\mathcal{D}_{\psi}$, $f_{\theta}$
\FOR{epoch $=1,\dots,E$}
  \FOR{mini-batch $B=\{(\bm{x}_n,\bm{y}_n)\}$}
    \STATE $(\bm{\mu}_n,\bm{\sigma}_n) \leftarrow \mathcal{E}_{\phi}(\bm{x}_n)$
    \STATE $ \bm{z}_n \leftarrow \bm{\mu}_n + \bm{\sigma}_n \odot \bm{\epsilon}_n, \quad \bm{\epsilon}_n \sim \mathcal N(\mathbf{0},\mathbf{I})$ \hfill $\triangleright$ reparam.
    \STATE $\hat{\bm{x}}_n \leftarrow \mathcal{D}_{\psi}(\bm{z}_n)$
    \STATE $\mathcal L_{\mathrm{VAE}} \leftarrow \frac{1}{|B|}\!\sum_{n\in B}\!\|\bm{x}_n-\hat{\bm{x}}_n\|_2^2
           + \beta\,\mathrm{KL}\!\big(\mathcal N(\bm{\mu}_n,\operatorname{diag}(\bm{\sigma}_n^2))\,\|\,\mathcal N(\mathbf{0},\mathbf{I})\big)$
    \STATE ${\bm{\hat{\mu}}}_n \leftarrow \textbf{\texttt{sg}}(\bm{\mu}_n)$ \hfill $\triangleright$ stop-grad
    \STATE $\bm{v} \leftarrow \big|\operatorname{corr}\big(\{{\bm{\hat{\mu}}}_n\}_{n\in B},\{\bm{y}_n\}_{n\in B}\big)\big|$ \hfill $\triangleright$ per-dimension 
    \STATE $\bar{\bm{z}}_n \leftarrow {\bm{z}}_n + \alpha\,\bm{v}\odot\bm{e}_n,  \quad \bm{e}_n \sim \mathcal N(\mathbf{0},\mathbf{I}) $
    \STATE $\mathcal L_{\mathrm{clf}} \leftarrow \frac{1}{|B|}\!\sum_{n\in B}\!
            \Big[\ell_{\mathrm{CE}}\!\big(f_{\theta}(\bar{\bm{z}}_n),\bm{y}_n\big)
            + \lambda\, \|f_{\theta}(\bar{\bm{z}}_n)-f_{\theta}({\bm{z}}_n)\|_2^2\Big]$
    \STATE $\mathcal L_{\mathrm{total}} \leftarrow \mathcal L_{\mathrm{VAE}} + \mathcal L_{\mathrm{clf}}$
    \STATE optimize $\phi,\psi,\theta$ to minimize $\mathcal L_{\mathrm{total}}$
  \ENDFOR
\ENDFOR
\STATE \textbf{Test-time:} given $\bm{x}$, compute $(\bm{\mu},\bm{\sigma})\!\leftarrow\!E_{\phi}(\bm{x})$ and return $\hat y=\arg\max f_{\theta}(\bm{\mu})$
\end{algorithmic}
\end{algorithm}

\section{SITAR as a Targeted Jacobian Regularizer}
\label{sec:theory}

Let $f_\theta\colon\mathbb{R}^m \to \mathbb{R}^C$ denote the classifier (logit) map and $\ell_{\text{CE}}(\cdot,y)$ the standard cross-entropy loss. Recall the anisotropic perturbation from Section~\ref{sec:method_step2},
\begin{align}
    \Delta\bm{z} = \alpha(\bm{v} \odot \bm{e}), \quad \bm{e} \sim \mathcal{N}(\bm{0}, I_m),  \nonumber\\   \Sigma \coloneqq \mathbb{E}[\Delta\bm{z} \Delta\bm{z}^\top] = \alpha^2 \operatorname{Diag}(\bm{v}^2).
\end{align}
We define the perturbed output $\bar{f} = f_\theta(\bm{z} + \Delta\bm{z})$ and clean output $f = f_\theta(\bm{z})$. Setting $\lambda_{\text{cons}}=1$ without loss of generality, the SITAR classifier objective is
\[
    \mathcal{L}_{\text{total}} = \mathbb{E}_{\bm{e}}\big[\ell_{\text{CE}}(\bar{f}, y) + \|\bar{f} - f\|_2^2\big].
\]

\begin{theorem}[Second-Order Expansion of SITAR]
\label{thm:sitar_expansion}
Assume $f_\theta$ and $\ell_{\text{CE}}$ are $C^3$-smooth in a neighborhood of $\bm{z}$. Let $g \coloneqq \nabla_f \ell_{\text{CE}}(f, y) \in \mathbb{R}^C$ be the gradient of the loss with respect to the logits, $H_\ell \coloneqq \nabla^2_f \ell_{\text{CE}}(f, y) \in \mathbb{R}^{C \times C}$ its Hessian, $J_f(\bm{z}) \in \mathbb{R}^{C \times m}$ the Jacobian of $f_\theta$ with respect to $\bm{z}$, and $H_{f_c}(\bm{z}) \in \mathbb{R}^{m \times m}$ the Hessian of the $c$-th logit with respect to $\bm{z}$. For sufficiently small $\alpha$,
\begin{align}
    \mathcal{L}_{\text{total}} = \ell_{\text{CE}}(f, y) 
    + \underbrace{\operatorname{tr}\!\left( \left(\tfrac{1}{2} H_\ell + I_C\right) J_f(\bm{z}) \Sigma J_f(\bm{z})^\top \right)}_{\text{Targeted Jacobian Regularization}} \nonumber \\
    + \underbrace{\tfrac{1}{2} \operatorname{tr}\!\left( \Sigma \sum_{c=1}^C g_c H_{f_c}(\bm{z}) \right)}_{\text{Curvature Regularization}} 
    + \mathcal{O}(\alpha^3). \label{eq:matrix_form}
\end{align}
Substituting $\Sigma = \alpha^2 \operatorname{Diag}(\bm{v}^2)$, this decomposes as

\begin{align}
    \mathcal{L}_{\text{total}} \approx \ell_{\text{CE}} + \alpha^2 \sum_{i=1}^m v_i^2 \Biggl( &\left\| J_{f,i}(\bm{z}) \right\|^2_{\frac{1}{2}H_\ell + I_C} \notag \\ 
    &+ \frac{1}{2} \sum_{c=1}^C g_c \frac{\partial^2 (f_\theta)_c}{\partial z_i^2} \Biggr).
    \label{eq:scalar_form}
\end{align}
\end{theorem}

\begin{proof}
Let $h(\bm{z}) \coloneqq \ell_{\text{CE}}(f_\theta(\bm{z}), y)$. The second-order Taylor expansion gives
\begin{align}
    \mathbb{E}[h(\bm{z} + \Delta\bm{z})] = h(\bm{z}) + \mathbb{E}[\nabla h^\top \Delta\bm{z}] \nonumber \\ + \tfrac{1}{2} \mathbb{E}[\Delta\bm{z}^\top \nabla^2 h(\bm{z}) \Delta\bm{z}] + \mathcal{O}(\alpha^3). \nonumber
\end{align}
Since $\mathbb{E}[\Delta\bm{z}]=\bm{0}$, the linear term vanishes. Using the identity $\mathbb{E}[\Delta\bm{z}^\top A\, \Delta\bm{z}] = \operatorname{tr}(A\Sigma)$, the quadratic term simplifies to $\frac{1}{2}\operatorname{tr}(\Sigma \nabla^2 h(\bm{z}))$. Applying the chain rule for Hessians,
\[
\nabla^2 h(\bm{z}) = J_f(\bm{z})^\top H_\ell\, J_f(\bm{z}) + \sum_{c=1}^C g_c H_{f_c}(\bm{z}),
\]
and using the cyclic property of the trace, we obtain
\begin{align*}
    \mathbb{E}[\ell_{\text{CE}}(\bar{f}, y)] \approx \ell_{\text{CE}}(f, y) + \tfrac{1}{2}\operatorname{tr}(H_\ell J_f(\bm{z}) \Sigma J_f(\bm{z})^\top) \\ + \tfrac{1}{2}\operatorname{tr}\!\left(\Sigma \sum_{c=1}^C g_c H_{f_c}(\bm{z})\right).
\end{align*}
For the consistency term, expanding $\bar{f} \approx f + J_f(\bm{z})\,\Delta\bm{z}$ to leading order,
\[
    \mathbb{E}[\|\bar{f} - f\|_2^2] = \operatorname{tr}(J_f(\bm{z})\,\Sigma\, J_f(\bm{z})^\top) + \mathcal{O}(\alpha^3).
\]
Summing both expectations and collecting terms yields Eq.~\eqref{eq:matrix_form}.
\end{proof}

\paragraph{Remark.}
Eq.~\eqref{eq:scalar_form} shows that SITAR implicitly applies both Jacobian and curvature regularization, with each penalty weighted by $v_i^2$, the squared label correlation of dimension $i$. Shortcut dimensions carry high $v_i$ and are therefore penalized most strongly, while core dimensions remain largely unaffected. 
When $f_\theta$ is locally affine, $H_{f_c}$ vanishes and SITAR reduces to a pure targeted Jacobian regularizer. 
This $v_i^2$-weighted anisotropy is the key mechanism distinguishing SITAR from standard uniform regularization, and provides a formal justification for its ability to suppress shortcut reliance without degrading performance on core features.

\begin{figure*}[t]
    \centering
    \begin{minipage}{0.48\linewidth}
        \centering
        \includegraphics[width=\linewidth]{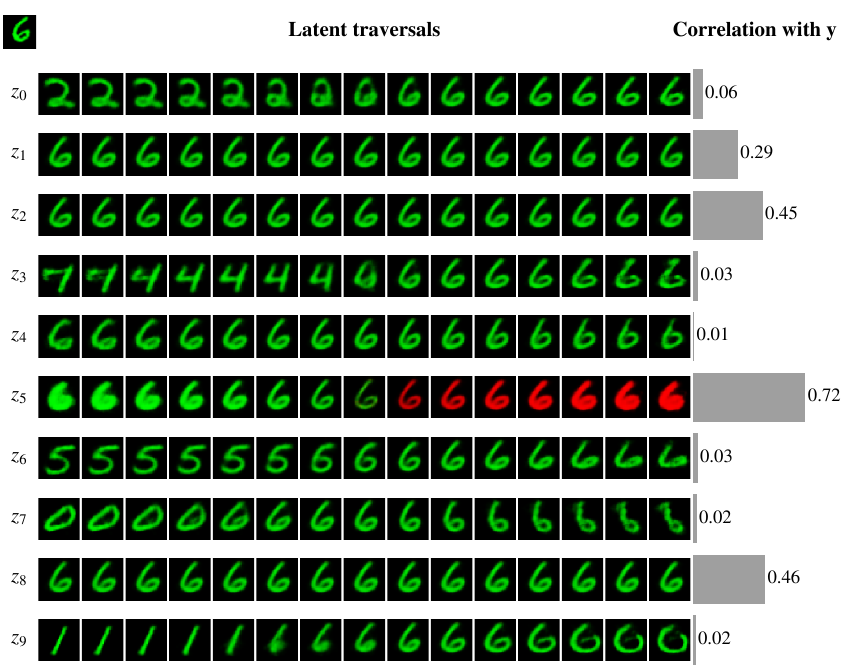}
    \end{minipage}
    \hfill
    \begin{minipage}{0.48\linewidth}
        \centering
        \includegraphics[width=\linewidth]{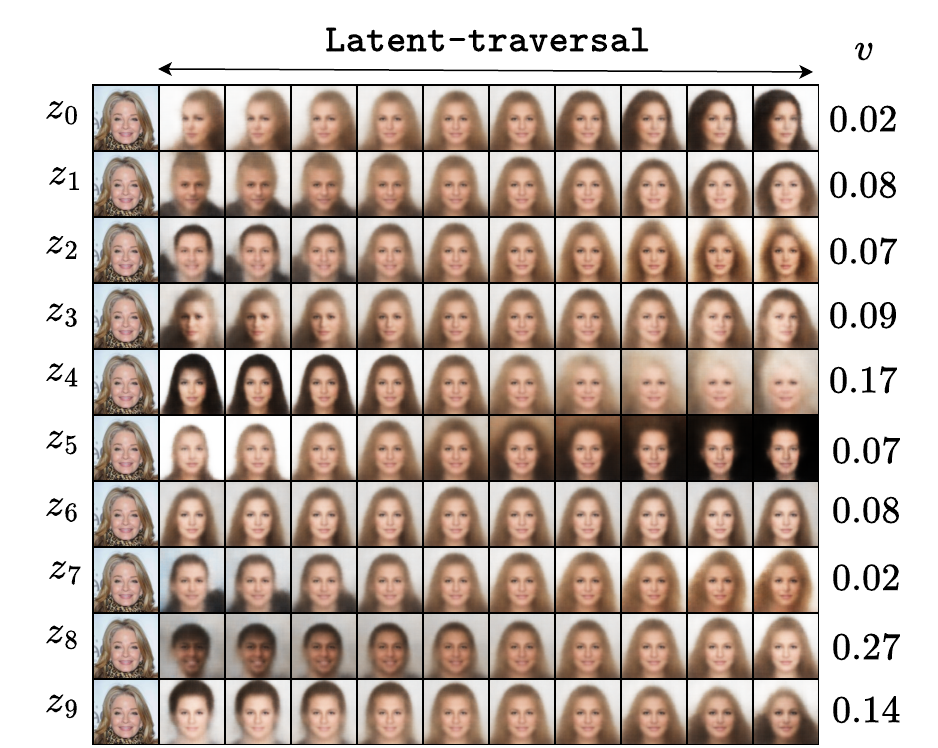}
    \end{minipage}
    \caption{Shortcut proxy $\bm{v}$ on \texttt{ColorMNIST} (left) and \texttt{CelebA} (right). Each row shows a latent traversal along a single dimension; the bar plot shows $v_j$ per dimension.}
    \label{fig:hypothesis_validation}
\end{figure*}

\section{Experiments}
We evaluate SITAR in three settings. We begin with \texttt{ColorMNIST} as a controlled environment to validate our core hypothesis, confirm that disentanglement is a necessary precondition, and ablate the key components of SITAR. We then evaluate on standard real-world shortcut learning benchmarks in both pixel space and on pretrained representations. Finally, we test on \texttt{Camelyon17-WILDS} to assess applicability to medical imaging and non-semantic shortcuts.

\subsection{Experimental Setup}
\label{sec:exp_setup}

\paragraph{ColorMNIST.} Following \cite{arjovsky2019invariant}, we construct \texttt{ColorMNIST} from \texttt{MNIST}. Digits 0--4 are mapped to label $y=0$ and digits 5--9 to $y=1$, then $y$ is independently flipped with probability $p_d=0.25$. Color is introduced by sampling a color label $c$ from $y$ with flip probability $p_c$. In the in-distribution split $\mathcal{D}_{\text{in}}$ we set $p_c=0.1$, creating a strong color-label correlation, and in the OOD split $\mathcal{D}_{\text{out}}$ we set $p_c=0.9$, reversing the correlation. Images are colored green when $c=0$ and red when $c=1$. A standard ERM classifier trained on $\mathcal{D}_{\text{in}}$ will exploit the $90\%$ color shortcut and fail on $\mathcal{D}_{\text{out}}$.

 \paragraph{Real-world benchmarks.} For pixel-space experiments we use \textbf{CelebA} with two tasks (Blond/Gender and Attractive/Smiling) and \textbf{Waterbirds}~\cite{wah2011caltech}, detailed in Section~\ref{sec:exp_real}. For pretrained representation experiments we additionally include \textbf{BAR}~\cite{nam2020learningfailuretrainingdebiased}, detailed in Section~\ref{sec:pretrained}. \textbf{Camelyon17-WILDS}~\cite{bandi2018detection} serves as a medical imaging benchmark and is detailed in Section~\ref{sec:exp_camelyon}.

\paragraph{Implementation.} Full implementation details including architectures, hyperparameters, and dataset details are provided in the appendix.

\subsection{Controlled Evaluation on \texttt{ColorMNIST}}
\label{sec:exp_cmnist}
We use \texttt{ColorMNIST} as a controlled testbed to systematically validate the three core components of SITAR. The shortcut (color) and target (digit shape) are clearly separable, making it possible to directly verify whether our correlation proxy identifies the correct latent dimension, whether disentanglement is necessary, and whether the targeted noise is the mechanism responsible for invariance.

\subsubsection{Hypothesis Validation}
\label{sec:exp_proxy}

We test whether shortcut axes are reliably identified by label correlation in a disentangled latent space. We train a $\beta$-VAE ($\beta=2$) on \texttt{ColorMNIST} without any label supervision and compute $v_j$ for each latent dimension over the training set. Figure~\ref{fig:hypothesis_validation} (left) shows the result. Each row displays a latent traversal obtained by varying a single dimension while keeping all others fixed, and the last column shows the bar plot of $v_j$ values. Dimension $z_5$ attains the highest correlation ($v_5 \approx 0.72$), and traversing along it changes digit color while largely preserving shape. Dimensions with low $v_j$ (e.g., $z_0, z_2, z_9$) modulate digit shape without affecting color.

We observe the same behavior on \texttt{CelebA} (Figure~\ref{fig:hypothesis_validation}, right). The first column shows the original images, the following columns show latent traversals, and the last column shows $|\mathrm{corr}(\bm{\mu}^{(j)}, y)|$ for each dimension. Dimension $z_8$ has a significantly higher correlation than all others, and traversing along it primarily changes apparent gender while leaving hair color largely unchanged. Images on the left appear more male-like and images on the right more female-like. Across both datasets, the correlation proxy $v_j$ consistently identifies the shortcut dimension without any label supervision, supporting our core hypothesis.

\begin{figure}
\vspace{-\intextsep}
\includegraphics[width=\linewidth]{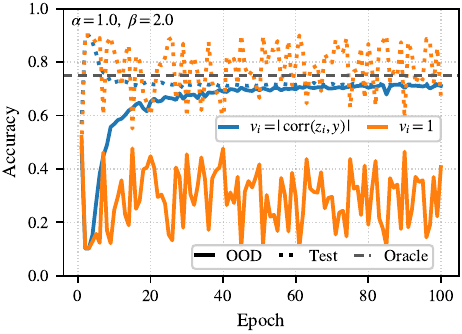}
\caption{Isotropic vs anisotropic noise}
\label{fig:isotropic_vs_anisotropic}
\end{figure}

\subsubsection{Disentanglement is a Necessary Precondition}
\label{sec:exp_beta}

\begin{figure*}
    \centering
    \begin{minipage}{0.48\linewidth}
        \centering
        \includegraphics[width=\linewidth]{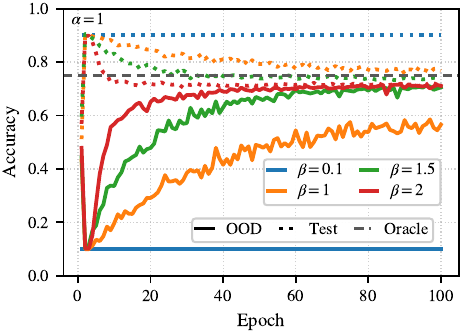}
    \end{minipage}
    \hfill
    \begin{minipage}{0.48\linewidth}
        \centering
        \includegraphics[width=\linewidth]{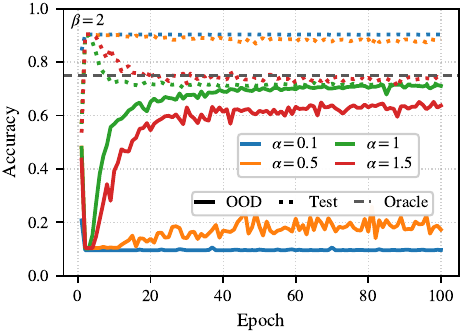}
    \end{minipage}
    \caption{Left: OOD accuracy versus disentanglement factor $\beta$ (fixed $\alpha=1.0$). Right: OOD accuracy versus noise magnitude $\alpha$ (fixed $\beta=2$).}
    \label{fig:ablations}
\end{figure*}

Our method relies on the $\beta$-VAE providing a sufficiently disentangled latent space. To verify this, we train SITAR with varying $\beta$ while fixing $\alpha=1.0$, and report OOD accuracy in Figure~\ref{fig:ablations} (left). For small $\beta$, the latent space is entangled and the correlation $v_j$ is dispersed across many dimensions rather than peaking on a single shortcut axis. In this regime SITAR is ineffective, and OOD accuracy collapses to ERM levels ($\sim\!10\%$). Once $\beta \geq 1$, the shortcut axis is isolated, $v_j$ peaks sharply on the color coordinate, the Jacobian penalty from Theorem~\ref{thm:sitar_expansion} concentrates on the correct dimension, and OOD accuracy rises toward the oracle ($\sim\!75\%$). This confirms that disentanglement is a necessary precondition for reliable shortcut identification and effective invariance.

\subsubsection{Targeted Noise is the Critical Mechanism}
\label{sec:exp_ablation}

We validate the two key components of the SITAR objective, the noise magnitude $\alpha$ and the targeting vector $\bm{v}$, through two separate ablations.


\noindent \textbf{Noise magnitude.} With a disentangled encoder ($\beta=2$), we sweep $\alpha$ and report OOD accuracy in Figure~\ref{fig:ablations} (right). \\
At $\alpha=0$, SITAR reduces to ERM and OOD accuracy collapses to $\sim\!10\%$. Increasing $\alpha$ consistently improves OOD performance, saturating around $65$--$70\%$. This is consistent with our theoretical analysis in Section~\ref{sec:theory}. Any non-zero $\alpha$ activates the targeted Jacobian penalty, shifting classifier sensitivity away from the high-$v_j$ shortcut axes.

\noindent \textbf{Noise targeting.} We compare our anisotropic noise ($\bm{v} = |\mathrm{corr}(\bm{\mu}, \bm{y})|$) against an isotropic baseline where $v_j = 1$ for all $j$. The isotropic baseline penalizes all dimensions equally, flattening useful and spurious directions alike. As shown in Figure~\ref{fig:isotropic_vs_anisotropic}, this untargeted baseline fails to improve OOD performance. Our anisotropic variant concentrates regularization on the high-$v_j$ dimensions, successfully flattens the classifier along the shortcut axis, and reaches near-oracle OOD accuracy. This confirms that targeting is the critical mechanism.

\subsection{Robustness Across Shortcut Severity}
\label{sec:exp_robustness}
\begin{table*}
\centering
\small
\caption{ID and OOD accuracy on \texttt{ColorMNIST} across shortcut severity $\rho$.}
\label{tab:performance_metrics}
\begin{tabular}{lcccccccc}
\toprule
\multirow{2}{*}{Methods} & \multicolumn{2}{c}{$\rho=0.5$} & \multicolumn{2}{c}{$\rho=0.7$} & \multicolumn{2}{c}{$\rho=0.9$} & \multicolumn{2}{c}{$\rho=1.0$} \\ 
\cmidrule(lr){2-3} \cmidrule(lr){4-5}\cmidrule(lr){6-7}\cmidrule(lr){8-9}
 & ID  & OOD  & ID  & OOD  & ID  & OOD  & ID  & OOD  \\ 
\midrule
ERM   & \bm{$74.4$} & \bm{$73.8$} & \bm{$73.8$} & 66.7 & \bm{$83.2$} & 27.3 & \bm{$100$} & 0.0 \\ 
JTT~\cite{liu2021justtraintwiceimproving}   & 55.3 & 56.3 & 63.2 & 53.6 & 74.7 & 61.2 & \bm{$100$} & 0.0\\ 
LfF~\cite{nam2020learningfailuretrainingdebiased}   & 59.8 & 58.5 & 68.5 & 64.3 & 69.1 & 69.6 & \bm{$100$} & 0.0 \\ 
\midrule
SITAR & 68.8 & 69.9 & 69.7 & \bm{$70.2$} & 72.1 & \bm{$71.5$} & 70.8  & \bm{$70.3$} \\ 
\bottomrule
\end{tabular}
\end{table*}

 Table~\ref{tab:performance_metrics} reports ID and OOD accuracies on \texttt{ColorMNIST} across varying levels of spurious correlation $\rho$, where $\rho$ denotes the fraction of training samples where color and label are aligned (shortcut aligned) or $1-\rho$ is fraction of shortcut conflicting samples. At $\rho=0.5$ color is uninformative and no shortcut exists, while at $\rho=1.0$ the training set has no shortcut-conflicting samples.
 
ERM and JTT perform competitively at $\rho=0.5$ but degrade sharply under stronger shortcuts, collapsing to $0\%$ OOD accuracy at $\rho=1.0$. LfF is more stable but similarly collapses at $\rho=1.0$. SITAR maintains consistent ID and OOD accuracy across all $\rho$, with OOD accuracy never dropping below $70\%$. Notably, all hyperparameters are fixed across $\rho$, confirming that this robustness is not the result of per-setting tuning.
This stability comes with a modest and expected trade-off. At $\rho=0.5$, where no shortcut is present, SITAR's ID accuracy ($69\%$) is slightly below ERM ($74\%$). This is consistent with the general principle that enforcing invariance constraints introduces a small cost in in-distribution performance when no shortcut is present to correct for.

A natural question is why SITAR does not degrade more severely at $\rho=0.5$, given that noise is still injected during training even when no shortcut exists. The key is the Functional Consistency term in Eq.~\eqref{eq:obj}, which requires $f_\theta(\bm{z}) \approx f_\theta(\bar{\bm{z}})$. This term does not suppress the perturbed dimensions entirely. It forces the classifier to remain consistent under the perturbation, meaning predictive signal in those dimensions is preserved as long as the classifier treats the clean and noisy inputs similarly. When no shortcut is present, $v_j$ values are more uniformly distributed and no single dimension receives disproportionately strong noise, so the Jacobian penalty is mild and spread evenly. The core signal therefore remains accessible to the classifier, and performance is largely maintained.

\section{Real-World Benchmarks}
\label{sec:exp_real}

We evaluate SITAR on standard shortcut learning benchmarks under a majority-only training regime. Specifically, we construct the training set to contain only the two majority, correlated groups, i.e., samples where the spurious attribute $S$ and label $Y$ are aligned. For example, in \textbf{Waterbirds} the training set contains only landbirds on land ($Y=0, S=0$) and waterbirds on water ($Y=1, S=1$), with the minority groups entirely withheld. This yields $\mathrm{corr}(Y, S \mid \mathcal{D}_{\text{train}}) = 1$, meaning the shortcut is a perfect predictor of the label during training and no shortcut-conflicting samples are available. Group labels are used only to construct the split and are never provided to any method during training.

We evaluate on three datasets. \textbf{CelebA}~\cite{liu2015faceattributes} is evaluated on two tasks: predicting blond hair (shortcut: gender) and predicting attractiveness (shortcut: smiling). \textbf{Waterbirds}~\cite{wah2011caltech} requires classifying bird type (waterbird vs.\ landbird) where the shortcut is the image background. We compare against \textbf{ERM}, \textbf{JTT}~\cite{liu2021justtraintwiceimproving}, \textbf{LfF}~\cite{nam2020learningfailuretrainingdebiased}, and \textbf{Chroma-VAE}~\cite{yang2022chroma}, the latter being the most direct competitor as it also uses a VAE but follows a robust representation paradigm. We report average accuracy (Avg) and worst-group accuracy (WG) as the primary metric of shortcut robustness.

\subsection{Pixel-Space Results}
\label{sec:exp_pixel}

\begin{table*}[t]
\small
\centering
\caption{Results on CelebA (Blond/Gender, Attractive/Smiling) and Waterbirds in pixel space. All models trained without shortcut-conflicting samples ($\mathrm{corr}(Y,S)=1$).}
\label{tab:pixel_results}
\begin{tabular}{lcccccc}
\toprule
\multirow{2}{*}{Method} & \multicolumn{2}{c}{CelebA (Blond/Gender)} & \multicolumn{2}{c}{CelebA (Attractive/Smiling)} & \multicolumn{2}{c}{Waterbirds} \\
\cmidrule(lr){2-3} \cmidrule(lr){4-5} \cmidrule(lr){6-7}
& Avg (\%) & WG (\%) & Avg (\%) & WG (\%) & Avg (\%) & WG (\%)\\
\midrule
ERM         & $64.20 \pm 2.44$ & $27.51 \pm 4.29$ & $54.42 \pm 0.22$ & $15.29 \pm 1.26$ & $56.81 \pm 0.93$ & $23.85 \pm 2.14$\\
JTT         & $64.69 \pm 1.48$ & $26.25 \pm 0.90$ & $54.36 \pm 0.56$ & $20.91 \pm 3.06$ & $62.27 \pm 9.25$ & $18.75 \pm 6.61$\\
LfF         & $62.10 \pm 0.86$ & $25.31 \pm 1.74$ & $53.01 \pm 1.99$ & $15.49 \pm 1.89$ & $57.76 \pm 3.49$ & $22.15 \pm 3.60$\\
Chroma-VAE  & $\bm{82.00 \pm 0.00}$ & $54.40 \pm 0.00$ & $\bm{66.90 \pm 0.00}$ & $53.10 \pm 0.00$ & $\bm{70.46 \pm 0.81}$ & $11.21 \pm 2.66$\\
\midrule
SITAR       & $81.13 \pm 1.30$ & $\bm{58.88 \pm 2.45}$ & $63.62 \pm 0.38$ & $\bm{60.95 \pm 1.67}$ & $57.13 \pm 0.33$ & $\bm{31.04 \pm 1.75}$\\
\bottomrule
\end{tabular}
\end{table*}

Table~\ref{tab:pixel_results} reports mean $\pm$ std over seeds. SITAR achieves the highest WG accuracy across all three benchmarks. On CelebA (Blond/Gender), SITAR reaches $58.88\%$ WG, outperforming Chroma-VAE by $+4.5$ points. On CelebA (Attractive/Smiling), SITAR achieves $60.95\%$ WG, exceeding Chroma-VAE by $+7.9$ points.

On Waterbirds, SITAR obtains $31.04\%$ WG, outperforming ERM, JTT, and LfF which remain in the $18$--$24\%$ range. The absolute WG accuracy is lower here than on CelebA, which we attribute to the complexity of the background shortcut. Unlike gender in CelebA, the background in Waterbirds is a high-dimensional, spatially distributed signal that is difficult to disentangle cleanly in pixel space. Consistent with this, the pretrained representation experiments in Section~\ref{sec:pretrained} show substantially stronger results on Waterbirds ($87.3\%$ WG), confirming that the limitation is in pixel-space disentanglement rather than in the SITAR objective itself.

Chroma-VAE is the strongest competitor on CelebA, where the shortcut is relatively localized and separable. However, it degrades severely on Waterbirds, dropping to $11.21\%$ WG, below ERM, which is consistent with our earlier argument that explicit latent partitioning is brittle when the shortcut signal is high-dimensional. SITAR, by contrast, regularizes classifier sensitivity along shortcut-aligned axes without attempting to partition the latent space, which remains effective even when disentanglement is partial.

\begin{figure}[t]
    \centering
    \includegraphics[width=0.99\linewidth]{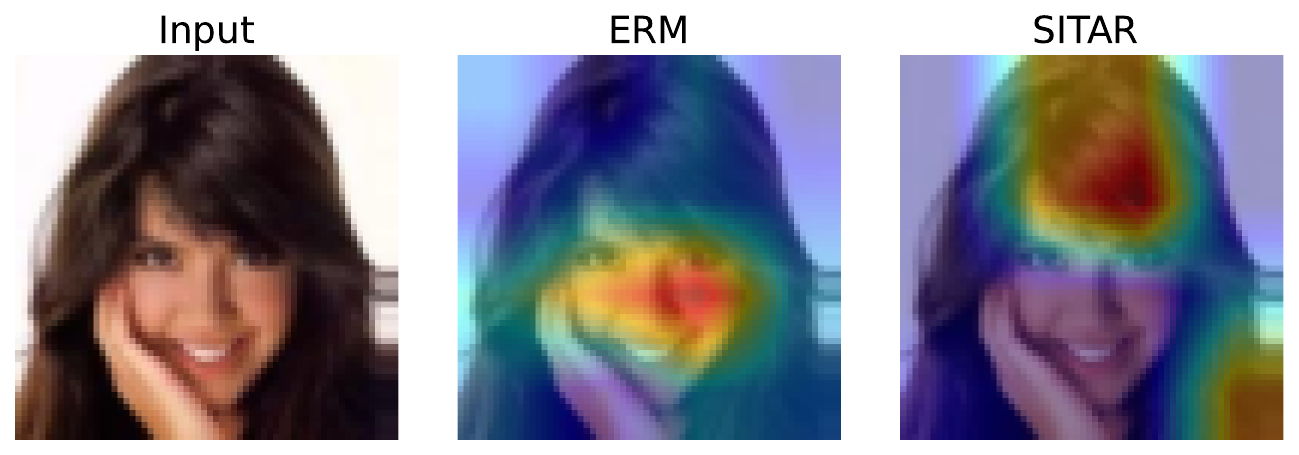}
    \caption{Grad-CAM visualizations on CelebA (Blond/Gender). ERM concentrates attention on the face region while SITAR shifts attention to the hair region.}
    \label{fig:gradcam_celeba}
\end{figure}

The Grad-CAM visualizations in Figure~\ref{fig:gradcam_celeba} provide qualitative support. ERM concentrates attention on the face region, which is informative for gender but only weakly related to hair color. SITAR shifts attention toward the hair region, which is directly relevant for the blond prediction, consistent with the quantitative results.

\subsection{Pretrained Representations}
\label{sec:pretrained}

Learning a disentangled representation directly in pixel space is fundamentally challenging for high-dimensional images~\cite{Locatello2019ChallengingCA}, and training a $\beta$-VAE on raw images is computationally costly due to the high-capacity decoder required. In this section we apply SITAR on top of pretrained representations instead. Concretely, we extract features using a frozen pretrained encoder (ResNet-50 for CelebA and Waterbirds, ResNet-18 for BAR) and train a $\beta$-VAE in this lower-dimensional feature space. Note that unlike the pixel-space experiments in Section~\ref{sec:exp_pixel}, these experiments use standard training splits that include shortcut-conflicting minority samples. Results are therefore not directly comparable across the two settings but demonstrate that SITAR scales to stronger encoders and more challenging benchmarks.

\begin{table*}[t]
\small
\centering
\caption{Results on CelebA, Waterbirds, and BAR using pretrained representations. Standard training splits are used.}
\label{tab:pretrained_results}
\begin{tabular}{lccccccc}
\toprule
\multirow{2}{*}{Method} & \multicolumn{2}{c}{CelebA} & \multicolumn{2}{c}{Waterbirds} & \multicolumn{1}{c}{BAR}\\
\cmidrule(lr){2-3} \cmidrule(lr){4-5} \cmidrule(lr){6-6}
& Avg (\%) & WG (\%) & Avg (\%) & WG (\%) & Avg (\%)\\
\midrule
ERM         & $\bm{95.6}$ & $47.2$ & $\bm{97.3}$ & $72.6$ & $60.5$ \\
JTT~\cite{liu2021justtraintwiceimproving}         & $88.0$ & $\bm{81.1}$ & $93.3$ & $86.7$ & $68.5$ \\
LfF~\cite{nam2020learningfailuretrainingdebiased}         & $85.1$ & $77.2$ & $91.2$ & $78.0$ & $62.9$ \\
Diffusion~\cite{li2025generative}   & $91.2$ & $69.4$ & $96.8$ & $79.4$ & $-$ \\
\midrule
SITAR       & $92.9$ & $\bm{81.1}$ & $93.1$ & $\bm{87.3}$ & $\bm{71.7}$ \\
\bottomrule
\end{tabular}
\end{table*}

Table~\ref{tab:pretrained_results} reports results across CelebA, Waterbirds, and BAR datasets. On CelebA, SITAR matches JTT on WG accuracy ($81.1\%$) while substantially improving over ERM ($47.2\%$), demonstrating that the correlation proxy successfully identifies the gender shortcut in the pretrained feature space. On Waterbirds, SITAR achieves the highest WG accuracy ($87.3\%$), outperforming JTT by $+0.6$ points and Diffusion by $+7.9$ points. On BAR, SITAR achieves the highest average accuracy ($71.7\%$), outperforming all baselines including JTT ($68.5\%$).

We fine-tune the pretrained encoder jointly during training. For all datasets we set the $\beta$-VAE latent dimensionality to $d/4$ where $d$ is the encoder output dimension, and use $\beta=2$ and $\alpha=0.1$ across all datasets without per-dataset tuning, which further demonstrates the robustness of SITAR to hyperparameter selection.

\section{Medical Imaging: Camelyon17-WILDS}
\label{sec:exp_camelyon}

We conduct a final evaluation on \textbf{Camelyon17-WILDS}~\cite{bandi2018detection} to test SITAR beyond natural images. This benchmark presents two distinct challenges relative to the previous experiments. The task is medical, requiring identification of tumor tissue from histopathology slides, a domain with fundamentally different low-level statistics from natural images. The shortcut is not a semantic object but a process-based domain artifact. Images are collected from five hospitals, and the hospital of origin is spuriously correlated with the label due to differences in staining protocols and scanner calibration. The standard split provides training data from three hospitals and evaluates on two unseen hospitals, making OOD accuracy the primary metric.

\begin{table}
\small
\vspace{-2pt}
\centering
\caption{Results on Camelyon17-WILDS. OOD accuracy is evaluated on two unseen hospitals.}
\label{tab:camelyon}
\begin{tabular}{lcc}
\toprule
\multirow{2}{*}{Method} & \multicolumn{2}{c}{Camelyon17-WILDS} \\
\cmidrule(lr){2-3}
& In-Dist (\%) & OOD (\%) \\
\midrule
ERM         & $\bm{95.59 \pm 0.08}$ & $81.63 \pm 0.63$ \\
JTT~\cite{liu2021justtraintwiceimproving}         & $94.71 \pm 0.71$      & $81.78 \pm 2.53$ \\
LfF~\cite{nam2020learningfailuretrainingdebiased}         & $95.40 \pm 0.25$      & $80.77 \pm 1.43$ \\
Chroma-VAE~\cite{yang2022chroma}  & $82.64 \pm 1.41$      & $74.45 \pm 0.65$ \\
\midrule
SITAR       & $95.41 \pm 0.61$      & $\bm{83.26 \pm 2.13}$ \\
\bottomrule
\end{tabular}
\end{table}

Table~\ref{tab:camelyon} reports in-distribution and OOD accuracy across methods. SITAR achieves the highest OOD accuracy ($83.26\%$), outperforming ERM by $+1.6$ points and JTT by $+1.5$ points, while maintaining competitive in-distribution accuracy ($95.41\%$). Chroma-VAE degrades substantially on this benchmark, dropping to $74.45\%$ OOD and $82.64\%$ in-distribution, suggesting that explicit latent partitioning struggles with the subtle, non-semantic nature of the hospital artifact shortcut.

The consistent improvement over ERM and reweighting baselines on a domain this different from natural images demonstrates that the correlation proxy $v_j$ is not limited to semantic or visually salient shortcuts. The subtle staining artifacts introduced by different hospitals manifest as structured variation in the latent space that is correlated with the label, and SITAR successfully identifies and suppresses this variation without any domain or group labels.
\section{Conclusion}
\label{sec:conclusion}

We propose SITAR, a simple latent-space framework for shortcut learning that enforces \emph{functional invariance} directly at the classifier level, without requiring shortcut labels, shortcut-conflicting samples, or explicit latent space partitioning. Our core insight is that in a disentangled representation, shortcut-aligned dimensions are reliably identified by their label correlation, providing an unsupervised proxy that requires no group supervision. Leveraging this signal, we introduce a targeted anisotropic noise objective that selectively perturbs shortcut-aligned axes during training, and show theoretically that this objective is equivalent to a targeted Jacobian regularizer that penalizes classifier sensitivity along shortcut directions weighted by their correlation strength.

Empirically, SITAR demonstrates strong and consistent worst-group accuracy across benchmarks spanning synthetic shortcuts, natural image shortcuts, and non-semantic domain artifacts in medical imaging. The pretrained representation experiments further show that SITAR scales naturally to high-dimensional settings where pixel-space disentanglement is insufficient, substantially broadening the applicability of functional robustness methods to real-world deployment.
{
    \small
    \bibliographystyle{ieeenat_fullname}
    \bibliography{main}

@String(CVPR= {IEEE Conf. Comput. Vis. Pattern Recog.})

@String(ICCV= {Int. Conf. Comput. Vis.})

@String(ECCV= {Eur. Conf. Comput. Vis.})

@String(ICLR = {Int. Conf. Learn. Represent.})

@String(CVPR  = {CVPR})

@String(ICCV  = {ICCV})

@String(ECCV  = {ECCV})

@String(ICLR  = {ICLR})

@book{vapnik1995,
author = {Vapnik, Vladimir N.},
title = {The nature of statistical learning theory},
year = {1995},
isbn = {0387945598},
publisher = {Springer-Verlag},
address = {Berlin, Heidelberg}
}

@article{geirhos2020shortcut,
   title={Shortcut learning in deep neural networks},
   volume={2},
   ISSN={2522-5839},
   url={http://dx.doi.org/10.1038/s42256-020-00257-z},
   DOI={10.1038/s42256-020-00257-z},
   number={11},
   journal={Nature Machine Intelligence},
   publisher={Springer Science and Business Media LLC},
   author={Geirhos, Robert and Jacobsen, Jörn-Henrik and Michaelis, Claudio and Zemel, Richard and Brendel, Wieland and Bethge, Matthias and Wichmann, Felix A.},
   year={2020},
   month=nov, pages={665–673} }

@misc{geirhos2018imagenet,
      title={ImageNet-trained CNNs are biased towards texture; increasing shape bias improves accuracy and robustness}, 
      author={Robert Geirhos and Patricia Rubisch and Claudio Michaelis and Matthias Bethge and Felix A. Wichmann and Wieland Brendel},
      year={2022},
      eprint={1811.12231},
      archivePrefix={arXiv},
      primaryClass={cs.CV},
      url={https://arxiv.org/abs/1811.12231}, 
}

@article{yang2022chroma,
  title={Chroma-vae: Mitigating shortcut learning with generative classifiers},
  author={Yang, Wanqian and Kirichenko, Polina and Goldblum, Micah and Wilson, Andrew G},
  journal={Advances in Neural Information Processing Systems},
  volume={35},
  pages={20351--20365},
  year={2022}
}

@inproceedings{
higgins2017beta,
title={beta-{VAE}: Learning Basic Visual Concepts with a Constrained Variational Framework},
author={Irina Higgins and Loic Matthey and Arka Pal and Christopher Burgess and Xavier Glorot and Matthew Botvinick and Shakir Mohamed and Alexander Lerchner},
booktitle={International Conference on Learning Representations},
year={2017},
url={https://openreview.net/forum?id=Sy2fzU9gl}
}

@misc{arjovsky2019invariant,
      title={Invariant Risk Minimization}, 
      author={Martin Arjovsky and Léon Bottou and Ishaan Gulrajani and David Lopez-Paz},
      year={2020},
      eprint={1907.02893},
      archivePrefix={arXiv},
      primaryClass={stat.ML},
      url={https://arxiv.org/abs/1907.02893}, 
}

@misc{sagawa2019distributionally,
      title={Distributionally Robust Neural Networks for Group Shifts: On the Importance of Regularization for Worst-Case Generalization}, 
      author={Shiori Sagawa and Pang Wei Koh and Tatsunori B. Hashimoto and Percy Liang},
      year={2020},
      eprint={1911.08731},
      archivePrefix={arXiv},
      primaryClass={cs.LG},
      url={https://arxiv.org/abs/1911.08731}, 
}

@misc{krueger2021out,
      title={Out-of-Distribution Generalization via Risk Extrapolation (REx)}, 
      author={David Krueger and Ethan Caballero and Joern-Henrik Jacobsen and Amy Zhang and Jonathan Binas and Dinghuai Zhang and Remi Le Priol and Aaron Courville},
      year={2021},
      eprint={2003.00688},
      archivePrefix={arXiv},
      primaryClass={cs.LG},
      url={https://arxiv.org/abs/2003.00688}, 
}

@techreport{wah2011caltech,
	Title = {Caltech-UCSD Birds-200-2011} ,
	Author = {Wah, C. and Branson, S. and Welinder, P. and Perona, P. and Belongie, S.},
	Year = {2011},
	Institution = {California Institute of Technology},
	Number = {CNS-TR-2011-001}
}

@misc{nam2020learningfailuretrainingdebiased,
      title={Learning from Failure: Training Debiased Classifier from Biased Classifier}, 
      author={Junhyun Nam and Hyuntak Cha and Sungsoo Ahn and Jaeho Lee and Jinwoo Shin},
      year={2020},
      eprint={2007.02561},
      archivePrefix={arXiv},
      primaryClass={cs.LG},
      url={https://arxiv.org/abs/2007.02561}, 
}

@inproceedings{liu2015faceattributes,
  title = {Deep Learning Face Attributes in the Wild},
  author = {Liu, Ziwei and Luo, Ping and Wang, Xiaogang and Tang, Xiaoou},
  booktitle = {Proceedings of International Conference on Computer Vision (ICCV)},
  month = {December},
  year = {2015} 
}

@misc{liu2021justtraintwiceimproving,
      title={Just Train Twice: Improving Group Robustness without Training Group Information}, 
      author={Evan Zheran Liu and Behzad Haghgoo and Annie S. Chen and Aditi Raghunathan and Pang Wei Koh and Shiori Sagawa and Percy Liang and Chelsea Finn},
      year={2021},
      eprint={2107.09044},
      archivePrefix={arXiv},
      primaryClass={cs.LG},
      url={https://arxiv.org/abs/2107.09044}, 
}

@article{bandi2018detection,
  title={From detection of individual metastases to classification of lymph node status at the patient level: the CAMELYON17 challenge},
  author={Bandi, Peter and Geessink, Oscar and Manson, Quirine and Van Dijk, Marcory and Balkenhol, Maschenka and Hermsen, Meyke and Bejnordi, Babak Ehteshami and Lee, Byungjae and Paeng, Kyunghyun and Zhong, Aoxiao and others},
  journal={IEEE Transactions on Medical Imaging},
  year={2018},
  publisher={IEEE}
}

@inproceedings{kim2018disentangling,
  title={Disentangling by Factorising},
  author={Kim, Hyunjik and Mnih, Andriy},
  booktitle={ICML},
  year={2018}
}

@article{muller2023shortcutvae,
  title={Shortcut Detection with Variational Autoencoders},
  author={Müller, Nicolas M. and Roschmann, Simon and Khan, Shahbaz and Sperl, Philip and Böttinger, Konstantin},
  journal={arXiv preprint arXiv:2302.04246},
  year={2023}
}

@article{scimeca2023,
  title={Mitigating Shortcut Learning with Diffusion Counterfactuals and Diverse Ensembles},
  author={Scimeca, Luca and Fan, Jianyu and Du, Mengting and Chen, Ching-Yao and Lee, Insu and Ribeiro, Marco Tulio},
  journal={arXiv preprint arXiv:2311.16176},
  year={2023}
}

@inproceedings{weng2024fastdiffusion,
  title={Fast Diffusion-Based Counterfactuals for Shortcut Removal and Generation},
  author={Weng, Zejiang and Zhang, Chengyue and Eftekhar, Armin and Zhang, Hong and Li, Ying},
  booktitle={ECCV},
  year={2024}
}

@inproceedings{li2025generative,
  title={Generative Classifiers Avoid Shortcut Solutions},
  author={Li, Yilun and Liu, Xinyu and Gu, Jiayi and Sinha, Aditya and Bottou, Léon and Lopez-Paz, David},
  booktitle={ICLR},
  year={2025}
}

@inproceedings{lim2023biasadv,
  author    = {Lim, Jongin and Kim, Youngdong and Kim, Byungjai and Ahn, Chanho and Shin, Jinwoo and Yang, Eunho and Han, Seungju},
  title     = {BiasAdv: Bias-Adversarial Augmentation for Model Debiasing},
  booktitle = {Proceedings of the IEEE/CVF Conference on Computer Vision and Pattern Recognition (CVPR)},
  year      = {2023},
  pages     = {3832-3841}
}

@article{morerio2024learnable,
  author  = {Morerio, Pietro and Ragonesi, Ruggero and Murino, Vittorio},
  title   = {Model Debiasing by Learnable Data Augmentation},
  journal = {arXiv preprint arXiv:2408.04955},
  year    = {2024}
}

@inproceedings{chang2021robust,
  title={Towards Robust Classification Model by Counterfactual and Invariant Data Generation},
  author={Chang, Ting-Yun and Lin, Chun-Hsiao and Yang, Wei and Li, Zhuoqing and Li, Hongliang and Hsu, Winston and Wen, Ying Nian and Chen, Tyng-Luh},
  booktitle={Proceedings of the IEEE/CVF Conference on Computer Vision and Pattern Recognition (CVPR)},
  pages={1301--1310},
  year={2021}
}

@inproceedings{Locatello2019ChallengingCA,
title	= {Challenging Common Assumptions in the Unsupervised Learning of Disentangled Representations},
author	= {Francesco Locatello and Stefan Bauer and Mario Lučić and Gunnar Rätsch and Sylvain Gelly and Bernhard Schölkopf and Olivier Frederic Bachem},
year	= {2019},
URL	= {http://proceedings.mlr.press/v97/locatello19a.html},
note	= {Best Paper Award},
booktitle	= {International Conference on Machine Learning}}

@InProceedings{Moosavi-Dezfooli_2019_CVPR,
author = {Moosavi-Dezfooli, Seyed-Mohsen and Fawzi, Alhussein and Uesato, Jonathan and Frossard, Pascal},
title = {Robustness via Curvature Regularization, and Vice Versa},
booktitle = {Proceedings of the IEEE/CVF Conference on Computer Vision and Pattern Recognition (CVPR)},
month = {June},
year = {2019}
}

@inproceedings{levy2020large,
  title={Large-Scale Methods for Distributionally Robust Optimization},
  author={Levy, Daniel and Carmon, Yair and Duchi, John C and Sidford, Aaron},
  booktitle={Advances in Neural Information Processing Systems},
  year={2020}
}

@article{DBLP:journals/corr/abs-1908-02729,
  publtype={informal},
  author={Judy Hoffman and Daniel A. Roberts and Sho Yaida},
  title={Robust Learning with Jacobian Regularization},
  year={2019},
  cdate={1546300800000},
  journal={CoRR},
  volume={abs/1908.02729},
  url={http://arxiv.org/abs/1908.02729}
}

@article{DeGrave2021,
  author    = {DeGrave, Alex J. and Janizek, Joseph D. and Lee, Su-In},
  title     = {AI for radiographic COVID-19 detection selects shortcuts over signal},
  journal   = {Nature Machine Intelligence},
  year      = {2021},
  volume    = {3},
  number    = {7},
  pages     = {610--619},
  month     = {jul},
  doi       = {10.1038/s42256-021-00338-7},
  url       = {https://doi.org/10.1038/s42256-021-00338-7},
  issn      = {2522-5839},
}

@misc{shortcut2023Nicolas,
      title={Shortcut Detection with Variational Autoencoders}, 
      author={Nicolas M. Müller and Simon Roschmann and Shahbaz Khan and Philip Sperl and Konstantin Böttinger},
      year={2023},
      eprint={2302.04246},
      archivePrefix={arXiv},
      primaryClass={cs.LG},
      url={https://arxiv.org/abs/2302.04246}, 
}
}

\clearpage
\maketitlesupplementary

\section{Remarks on the Theorem \ref{thm:sitar_expansion}}
\label{app:theory}

\paragraph{Decomposition of the Jacobian penalty.}
The matrix $\frac{1}{2}H_\ell + I_C$ combines contributions from the two objective terms. The $\frac{1}{2}H_\ell$ term arises from the robust CE loss and penalizes the component of the Jacobian that lies in the direction of high loss curvature, while the $I_C$ term arises from the consistency loss and penalizes the full Jacobian norm uniformly across output dimensions. Together they impose a stronger penalty on Jacobian columns that are both large in norm and aligned with directions of high loss sensitivity.

\paragraph{Curvature regularization.}
The second regularizer, $\frac{\alpha^2}{2}\sum_i v_i^2 \sum_c g_c 
\frac{\partial^2 (f_\theta)_c}{\partial z_i^2}$, penalizes the diagonal of the 
Hessian of each logit, weighted by the gradient of the loss. For ReLU-based 
classifiers $H_{f_c} = 0$ almost everywhere, which would eliminate this term 
uniformly regardless of $v_i^2$, undermining the targeted nature of the 
regularization. We therefore use a GELU-based classifier, for which $H_{f_c} \neq 0$, 
ensuring the curvature penalty remains active and anisotropic, concentrated on 
shortcut-aligned dimensions with high $v_i^2$.

\paragraph{Role of $\alpha$.}
The strength of both regularizers scales as $\alpha^2$. At $\alpha = 0$ the objective reduces to standard ERM. For any $\alpha > 0$ the targeted penalty is active, which explains the empirical observation that even small non-zero $\alpha$ yields OOD improvement (Section~\ref{sec:exp_ablation}).

\paragraph{Non-Gaussianity.}
The result holds for any zero-mean perturbation with
$\mathbb{E}[\Delta z_i \Delta z_j] = \alpha^2 v_i^2 \delta_{ij}$. Gaussianity of $\bm{e}$ is not required for the second-order analysis; it is used only to ensure the $\mathcal{O}(\alpha^3)$ remainder is well-controlled via bounded third-order derivatives of $h$.

\section{Comparison with Baseline Methods}
\label{app:baselines}

\textbf{LfF}~\cite{nam2020learningfailuretrainingdebiased} trains a biased model 
alongside the main model and upweights samples the biased model finds difficult, 
treating them as shortcut-conflicting. \\
\textbf{JTT}~\cite{liu2021justtraintwiceimproving} 
follows a similar philosophy. It identifies misclassified samples from a first-pass 
ERM model and upweights them in a second training run. Both methods operate in input 
space and implicitly require shortcut-conflicting samples to generate a meaningful 
reweighting signal. \\
\textbf{Chroma-VAE}~\cite{yang2022chroma} trains a $\beta$-VAE 
with a partitioned latent space where a small compressed subspace is used to train 
a classifier with standard ERM. Since shortcuts are the easiest features to learn, 
ERM naturally routes them into this compressed partition. The remaining subspace is 
then assumed to be shortcut-free and a final classifier is trained on it offline. 
This relies on the shortcut being strongly present so that ERM reliably separates it, 
and fails when the shortcut is absent or entangled with semantic content. \\
\textbf{Diffusion}~\cite{li2025generative} uses class-conditional diffusion models 
for classification via likelihood scoring, where robustness arises from the generative 
model being forced to model all features rather than the easiest discriminative signal. 
However, training a separate diffusion model per class is computationally expensive 
compared to discriminative approaches.

SITAR differs from all of the above in that it neither reweights samples nor removes 
shortcut information from the representation. Instead it preserves the full latent 
space and directly regularizes the classifier's sensitivity along shortcut-aligned 
dimensions, requiring no shortcut labels, no conflicting samples, and no explicit 
latent partitioning.
\section{Pixel-Space Experiments}

\subsection{Datasets}
For all pixel-space experiments (CelebA, Waterbirds, and Camelyon17), images are 
resized to $64 \times 64$ resolution. No data augmentation is applied. All baseline 
methods are trained under the same preprocessing pipeline and resolution to ensure 
a fair comparison.

\textbf{CelebA (Blond Hair / Gender)}
The target label is \textit{Blond Hair} (Not Blond vs.\ Blond) and the shortcut 
label is \textit{Gender} (Female vs.\ Male). The train and validation splits contain 
only the two majority groups, Not Blond Male and Blond Female, while the OOD test 
split contains all four group combinations. Table~\ref{tab:celeba_blond_splits} 
reports the exact group counts.

\begin{table}[h]
\centering
\small
\setlength{\tabcolsep}{4pt}
\caption{CelebA Blond Hair / Gender split statistics.}
\label{tab:celeba_blond_splits}
\begin{tabular}{llrrr}
\toprule
Split & Target & \multicolumn{2}{c}{Group counts} & Class count \\
\cmidrule(lr){3-4}
 & & Female & Male & \\
\midrule
\multirow{2}{*}{Train}
 & Not blond & 0        & 66{,}874 & 66{,}874 \\
 & Blond     & 22{,}880 & 0        & 22{,}880 \\
\midrule
\multirow{2}{*}{Val}
 & Not blond & 0       & 8{,}276 & 8{,}276 \\
 & Blond     & 2{,}874 & 0       & 2{,}874 \\
\midrule
\multirow{2}{*}{Test (OOD)}
 & Not blond & 9{,}767 & 7{,}535 & 17{,}302 \\
 & Blond     & 2{,}480 & 180     & 2{,}660 \\
\bottomrule
\end{tabular}
\end{table}

\textbf{CelebA (Attractive / Smiling)}
The target label is \textit{Attractive} (Not Attractive vs.\ Attractive) and the 
shortcut label is \textit{Smiling} (Not Smiling vs.\ Smiling). The train and 
validation splits contain only the two majority groups, Not Attractive Smiling and 
Attractive Not Smiling, while the OOD test split contains all four combinations. 
Table~\ref{tab:celeba_attr_splits} reports the exact group counts.

\begin{table}[h]
\centering
\small
\setlength{\tabcolsep}{3pt}
\caption{CelebA Attractive / Smiling split statistics.}
\label{tab:celeba_attr_splits}
\begin{tabular}{llrrr}
\toprule
Split & Target & \multicolumn{2}{c}{Group counts} & Class count \\
\cmidrule(lr){3-4}
 & & Not smiling & Smiling & \\
\midrule
\multirow{2}{*}{Train}
 & Not attractive & 0        & 31{,}956 & 31{,}956 \\
 & Attractive     & 37{,}479 & 0        & 37{,}479 \\
\midrule
\multirow{2}{*}{Val}
 & Not attractive & 0       & 3{,}780 & 3{,}780 \\
 & Attractive     & 4{,}510 & 0       & 4{,}510 \\
\midrule
\multirow{2}{*}{Test (OOD)}
 & Not attractive & 5{,}668 & 4{,}396 & 10{,}064 \\
 & Attractive     & 4{,}307 & 5{,}591 & 9{,}898 \\
\bottomrule
\end{tabular}
\end{table}

\textbf{Waterbirds}
The target label is bird type (Landbird vs.\ Waterbird) and the shortcut label is 
background (Land vs.\ Water). Train and validation splits contain only the two 
majority groups. The OOD test split contains all four groups. 
Table~\ref{tab:waterbirds_splits} reports the exact group counts.

\begin{table}[h]
\centering
\small
\setlength{\tabcolsep}{4pt}
\caption{Waterbirds split statistics.}
\label{tab:waterbirds_splits}
\begin{tabular}{llrrr}
\toprule
Split & Target & \multicolumn{2}{c}{Group counts} & Class count \\
\cmidrule(lr){3-4}
 & & Land & Water & \\
\midrule
\multirow{2}{*}{Train}
 & Landbird  & 3{,}498 & 0       & 3{,}498 \\
 & Waterbird & 0       & 1{,}057 & 1{,}057 \\
\midrule
\multirow{2}{*}{Val}
 & Landbird  & 467 & 0   & 467 \\
 & Waterbird & 0   & 133 & 133 \\
\midrule
\multirow{2}{*}{Test (OOD)}
 & Landbird  & 2{,}255 & 2{,}255 & 4{,}510 \\
 & Waterbird & 642     & 642     & 1{,}284 \\
\bottomrule
\end{tabular}
\end{table}

\textbf{Camelyon17-WILDS}
The task is binary classification of tumor tissue from histopathology slides. Images 
are collected from five hospitals; the standard WILDS split~\cite{bandi2018detection} 
uses three hospitals for training and two unseen hospitals for OOD evaluation. The 
hospital of origin serves as an implicit shortcut due to systematic differences in 
staining protocols and scanner calibration across sites. No group or hospital labels 
are provided to any method during training.

\subsection{Implementation Details}

\paragraph{$\beta$-VAE.}
We use a convolutional $\beta$-VAE with latent dimension $d$ across all pixel-space 
datasets. The encoder takes a $3 \times 64 \times 64$ RGB image and applies four 
convolutional layers with kernel size $4$, stride $2$, and padding $1$, each followed 
by ReLU activation, progressively expanding the channel dimension as 
$3 \to 32 \to 32 \to 64 \to 128$. This reduces the spatial resolution from 
$64 \times 64$ to $4 \times 4$, producing a $4 \times 4 \times 128 = 2048$ 
dimensional feature vector. Two separate MLP heads project this vector to the 
posterior mean $\bm{\mu} \in \mathbb{R}^d$ and log-variance 
$\log \bm{\sigma}^2 \in \mathbb{R}^d$. The decoder mirrors the encoder with four 
transposed convolutional layers. The reconstruction loss is per-pixel mean squared 
error and the KL term uses the standard closed-form expression for Gaussian posteriors. 
For \texttt{ColorMNIST}, a simpler two-layer convolutional encoder and decoder is used 
given the lower complexity of the data.

\paragraph{Classifier.}
A small MLP classifier is attached to the latent representation. It consists of a 
fully connected layer with 128 hidden units and GELU activation, followed by a linear 
output layer producing $C$ logits.

\paragraph{Correlation proxy.}
The shortcut sensitivity vector $\bm{v}$ is computed batchwise from the detached 
posterior means $\texttt{sg}(\bm{\mu})$ using the absolute Pearson correlation between 
each latent dimension and the label. The implementation is given below.

\begin{lstlisting}[style=pythonstyle]
def corr(X, y, eps=1e-8):
    # Center features and label
    Xc = X - X.mean(dim=0, keepdim=True)
    yc = y - y.mean()

    # Covariance and variances
    cov_xy = (Xc * yc[:, None]).mean(dim=0)
    var_x  = (Xc ** 2).mean(dim=0)
    var_y  = (yc ** 2).mean()

    # Pearson correlation per dimension
    r = cov_xy / (var_x.clamp_min(eps).sqrt() *
                  (var_y + eps).sqrt())
    return r.abs()
\end{lstlisting}

\section{Pretrained Representation Experiments}

We additionally evaluate on the Biased Action Recognition (BAR) 
dataset~\cite{nam2020learningfailuretrainingdebiased}, which contains 6-class action 
recognition images where each action class is spuriously correlated with a background 
scene during training. To ensure a fair comparison, we adopt the same standard dataset 
splits and evaluation protocol used by the baseline methods, and report results 
directly from the respective papers.

\subsection{Implementation Details}

For CelebA and Waterbirds we use a ResNet-50 pretrained on ImageNet, for BAR we use 
a ResNet-18. Images are resized to $224 \times 224$ and standard data augmentation 
is applied. The ResNet encoder is fine-tuned jointly with the $\beta$-VAE and 
classifier throughout training.

\paragraph{$\beta$-VAE in feature space.}
Rather than a convolutional architecture, we use a lightweight MLP-based $\beta$-VAE 
operating directly on the ResNet feature vector. The encoder consists of two fully 
connected layers with GELU activation, projecting from $d_{\text{enc}}$ to 
$d_{\text{enc}}/2$ and then to $d_{\text{enc}}/4$, with separate heads for the 
posterior mean and log-variance. The decoder mirrors this with two fully connected 
layers projecting from $d_{\text{enc}}/4$ back to $d_{\text{enc}}$. The latent 
dimension is set to $d_{\text{enc}}/4$, i.e., $512$ for ResNet-50 and $128$ for 
ResNet-18.

\subsection{Training Procedure}
All models are trained end-to-end with Adam using learning rate 
$10^{-4}$ and weight decay $10^{-2}$, with batch size $128$. The VAE encoder 
$\mathcal{E}_\phi$ receives gradients from both $\mathcal{L}_{\text{VAE}}$ and 
$\mathcal{L}_{\text{clf}}$, allowing the representation to be jointly shaped by both 
the reconstruction and classification objectives. The correlation vector $\bm{v}$ is 
computed from detached latent means at each step and treated as fixed when 
parameterizing the noise. The expectations in Eq.~\eqref{eq:obj} are approximated 
with a single-sample Monte Carlo estimate per forward pass. Model selection follows 
the same protocol as the baseline methods, using worst-group accuracy on a held-out 
validation set.

\section{Ablation on $\lambda_{\text{cons}}$}

\begin{figure}
\centering
\includegraphics[width=1.0\linewidth]{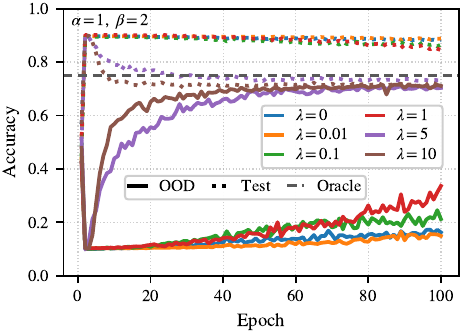}
\caption{OOD and in-distribution accuracy on \texttt{ColorMNIST} as a function of $\lambda_{\text{cons}}$ (fixed $\alpha=1.0$, $\beta=2.0$).}
\label{fig:lambda_ablation}
\end{figure}

Figure~\ref{fig:lambda_ablation} reports OOD and in-distribution accuracy on \texttt{ColorMNIST} as a function of $\lambda_{\text{cons}} \in \{0, 0.01, 0.1, 1, 5, 10\}$, with $\alpha=1.0$ and $\beta=2$ fixed. At $\lambda_{\text{cons}}=0$ the consistency term is absent and the method reduces to ERM despite non-zero $\alpha$ and $\beta$, confirming that the consistency objective is the active mechanism for shortcut suppression. Small values ($0.01$, $0.1$) yield modest gains, while larger values ($1$, $5$, $10$) produce substantially higher OOD accuracy approaching oracle performance. This is consistent with Theorem~\ref{thm:sitar_expansion}: increasing $\lambda_{\text{cons}}$ strengthens the induced Jacobian regularization, pushing the classifier further toward shortcut-invariant solutions.

\end{document}